%% file: root.tex
\documentclass{ifacconf}

\usepackage{graphicx}      
\usepackage{natbib}        

\usepackage{amsmath} 
\usepackage{amssymb}  
\usepackage{graphicx}

\usepackage{bm}

\input{commands}
\begin{document}
\begin{frontmatter}

\title{Tensor Decompositions for Modeling Inverse Dynamics} 

\author[First]{Stephan Baier} 
\author[Second]{Volker Tresp} 

\address[First]{Ludwig Maximilian University of Munich, 
   Oettingenstr. 67, 80538 Munich (e-mail: stephan.baier@campus.lmu.de).}
\address[Second]{Siemens AG and Ludwig Maximilian University of Munich, 
   Otto-Hahn-Ring 6, 81739 Munich (e-mail: volker.tresp@siemens.com)}

\begin{abstract}                

Modeling inverse dynamics is crucial for accurate feedforward robot control. The model computes the necessary joint torques, to perform a desired movement. The highly non-linear inverse function of the dynamical system can be approximated using regression techniques. We propose as regression method a tensor decomposition model that exploits the inherent three-way interaction of \textit{positions} $\times$ \textit{velocities} $\times$ \textit{accelerations}. Most work in tensor factorization has addressed the decomposition of dense tensors. In this paper, we build upon the decomposition of sparse tensors, with only small amounts of nonzero entries. The decomposition of sparse tensors has successfully  been used in relational learning, e.g., the modeling of large knowledge graphs. Recently, the approach has been extended to multi-class classification with discrete input variables. Representing the data in high dimensional sparse tensors enables the approximation of complex highly non-linear functions. In this paper we show how the decomposition of sparse tensors can be applied to regression problems. Furthermore, we extend the method to continuous inputs, by learning a mapping from the continuous inputs to the latent representations of the tensor decomposition, using basis functions. We evaluate our proposed model on a dataset with trajectories from a seven degrees of freedom SARCOS robot arm. Our experimental results show superior performance of the proposed functional tensor model, compared to challenging state-of-the art methods. 

 

%

%

\end{abstract}


\end{frontmatter}

\section{Introduction}

Within model-based robot control, an inverse dynamics model is used to compute the necessary joint torques of the robot's motors for the execution of a desired movement. The feedforward control command can be calculated using the rigid-body formulation $u_{FF} = M(q) \ddot{q} + F(q, \dot{q})$, with $q, \dot{q}, \ddot{q}$ being vectors of joint positions, joint velocities, and joint accelerations. However, in practice many nonlinearities such as friction or actuator forces need to be taken into account. Thus, methods modeling  $ u_{FF} = f(q, \dot{q}, \ddot{q})$ using non-linear regression techniques have shown superior performance in inferring the required joint torques for feedforward robot control. The parameters of the function $f$ are estimated offline using collected trajectories of the robot. \cite{craig, schoelkopf, schaal}.


Tensor models have been applied successfully in many application areas, e.g., relational learning, multilinear time invariant systems,  factor analysis, and spatio-temporal analysis, see \cite{rescal, lichtenberg, morup, spatio-temporal}. Most literature on tensor modeling, however, is concerned with the decomposition of dense tensors, i.e., most of the elements in the tensor are nonzero. Models for sparse tensors have mainly become popular for the application of modeling large knowledge graphs, such as Yago, DBpedia, and Freebase, see \cite{rescal, yago, dbpedia}. In these models, the elements of the tensor represent all possible triple combinations of entities and relations in the knowledge graph. Only elements that represent known facts from the knowledge graph are set to one. This results in a very sparse tensor, where the vast majority of elements are zero. Recently, the approach has been extended to higher order tensors for the task of classifying discrete sensor data, see \cite{mfi}. The tensor represents the space of all possible combinations of sensor values. By learning a representation for each possible value of all sensors, the decomposition allows for approximating highly non-linear functions. 

In this paper we build upon the approach of decomposing sparse tensors, and apply it to inverse system identification. Our model exploits the inherent three-way interaction of \textit{positions} $\times$ \textit{velocities} $\times$ \textit{accelerations}. We first show how the method can be applied to regression tasks.  Furthermore, we extend the approach to continuous inputs, by including basis functions that map the continuous inputs to the latent representations of the tensor decompositions. In this way, we retrieve a functional version of tensor decompositions. The basis functions also imply smoothness on the inputs, such that the model is able to generalize well, in spite of the extreme sparsity. By using multivariate basis functions we can group inputs, such that the dimensionality of the tensor decomposition can be reduced. In our inverse dynamics model we group the joint positions, velocities, and accelerations of all degrees of freedom of the robot, resulting in a tensor of order three. This makes the powerful Tucker decomposition applicable to the problem.


We evaluate our model on a dataset of a seven degrees of freedom SARCOS robot arm that was introduced in \cite{lwpr}. An inverse dynamics model is learned based on collected trajectories, and its performance is evaluated on a 10 percent test set. The results show that our model outperforms a number of competitive baseline methods, such as linear regression, radial basis function networks (RBF-networks), and support vector regression. Furthermore, the Tucker model shows superior performance over a PARAFAC model.

The paper is structured as follows. The next section gives an overview of related work. Section \ref{basis_functions} shows how the factorization of sparse tensors can be utilized for regression problems, and how the tensor decompositions can be extended to continuous inputs, using basis functions. In Section \ref{inverse_dynamics_model} we describe a functional Tucker decomposition for the task of modeling inverse dynamics. Related work is discussed in Section \ref{related_work}. Section \ref{experiments} presents the experimental evaluation. Finally, we conclude our work in Section \ref{conclusion}.

\begin{figure*}[!t]
	\scalebox{1}{
		\includegraphics[width=\textwidth]{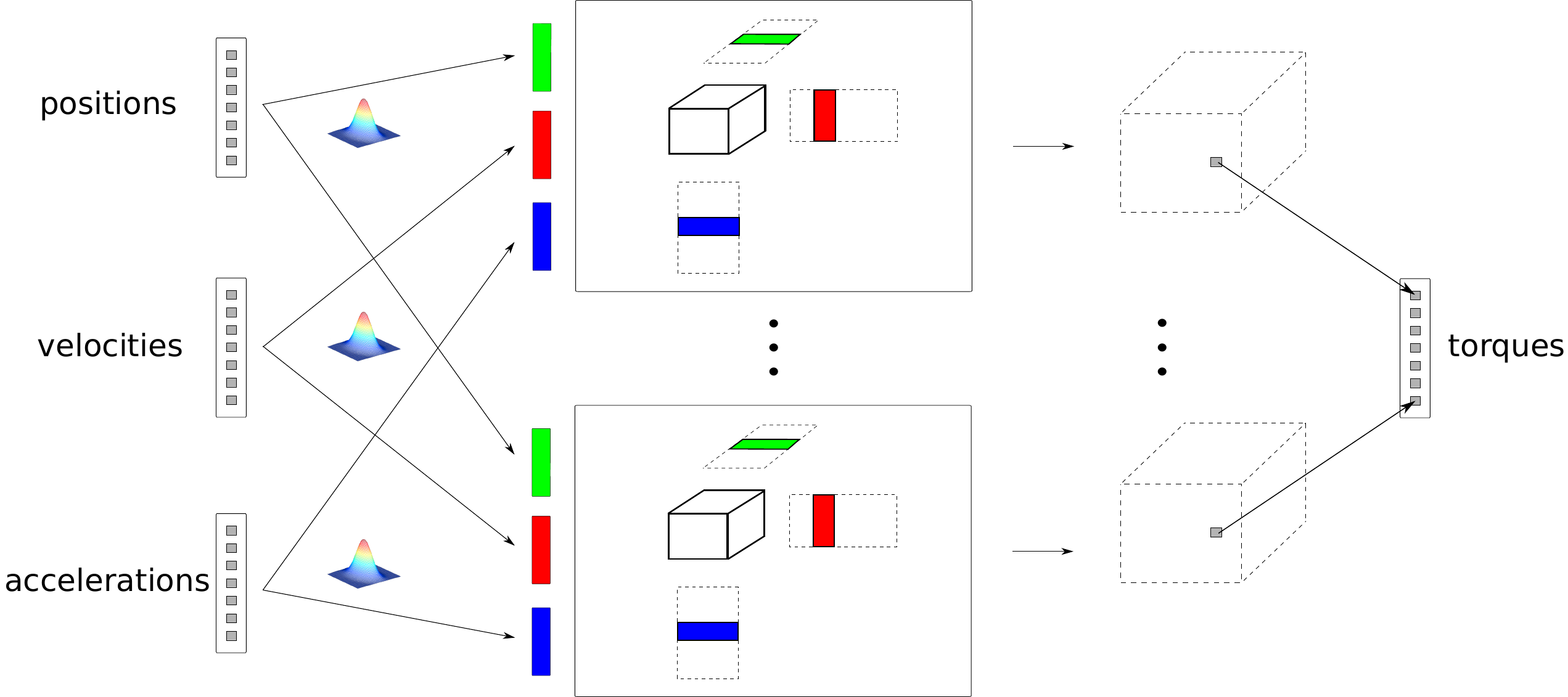}
	}
	\caption{Inverse dynamics model using a functional Tucker decomposition. The output tensors and the representation matrices are replaced by functions (illustrated with dashed lines). The representations are computed given the continuous inputs using Gaussian kernels.}
	\label{model}
\end{figure*}

\section{Tensor Decompositions using Basis Functions}
\label{basis_functions}
In this section we first show how the decomposition of sparse tensors can be applied to regression problems with discrete input variables. We then extend the model to continuous inputs, by using basis functions, which map the continuous input to the latent representations of the tensor decompositions. 

\subsection{Tensor Decompositions}

Tensor decompositions are a generalization of low rank matrix factorizations to higher order tensors. There are multiple ways of decomposing a higher order tensor. 

The full Tucker decomposition factorizes a tensor $\tensor{Y} \in \setR^{d_1 \times \dots \times d_S} $ into $S$ matrices, including latent representations for all fibers in each mode. The tensor elements are expressed by the interaction of the latent representations, weighted by a core tensor $\tensor{G} \in \setR^{\tilde{r} \times \dots \times \tilde{r}}$ such that
\begin{equation}
\begin{split}
\tensor{Y}(v_1, \dots, v_S) \approx \sum_{r_1, \dots, r_S}^{\tilde{r}} \tensor{G}(r_1, \dots, r_S) \cdot A_1(v_1, r_1) \cdot \\ A_2(v_2, r_2) \cdot \ldots \cdot A_{S}(v_{S}, r_S)
\end{split}
\label{tucker}
\end{equation}
with $A_i \in \setR^{d_i \times \tilde{r}}$. The full Tucker decomposition does not scale to high dimensions, as the core tensor $\tensor{G}$ grows exponentially with the dimensionality of the tensor; see \cite{Tucker}.

A special case of the full Tucker decomposition is the PARAFAC decomposition, where the core tensor $\tensor{G}$ is diagonal. All other interactions are left out, such that
\begin{equation}
\tensor{Y}(v_1, v_2,...,v_{S}) \approx \sum_{r=1}^{\tilde{r}} g(r) \cdot A_1(v_1, r) \cdot A_2(v_2, r)\cdot \ldots \cdot A_{S}(v_{S}, r).
\label{parafac}
\end{equation}
with $g \in \setR^{\tilde{r}}$. As PARAFAC only models the diagonal of the core tensor, its parameters scale linearly with the order of the tensor; see \cite{PARAFAC}.

\subsection{Discrete Input Regression}

We consider a regression problem with $S \in \setN$ discrete input variables. Each of the input variables $v_i$ for $i \in \{ 1,...,S \}$ assumes one out of $F_i \in \setN $ discrete values. Furthermore, we consider a dependent variable $y$. We model a regression function for a dataset of $N$ training examples $\{y^j, (v_1^j, ..., v_S^j)\}_{j=1}^N$.

All training examples are mapped to a sparse tensor $\tensor{Y} \in \setR^{F_1, ..., F_S}$. The tensor is filled with
\begin{equation}
\tensor{Y}(v_1^j, ..., v_S^j) = y^j \hspace{1em} \forall j \in \{1,...,N\}.
\end{equation}
The remaining entries of the tensor, which do not occur in the training data, are left unknown. This results in $\tensor{Y}$ being a very sparse tensor.

The tensor $\tensor{Y}$ is approximated using a low-rank tensor decomposition, e.g., the PARAFAC decomposition see equation \ref{parafac}. Using low ranks for $\tilde{r}$, the approximation results in a dense tensor $\Phi$. It describes the outcome $y$ for all combinations of the input variables $(v_1, ..., v_S)$. However, it would be impossible to compute and store the whole approximated tensor $\Phi$; thus, only the parameters of the decomposition are stored. When predicting $y$ for a new set of input variables, the representations for that tuple are indexed, and the approximation is computed on demand. In principle any tensor decomposition can be used for the approximation. However, in practice only few decompositions, such as PARAFAC and Tensor Train are scale-able to many dimensions, see \cite{PARAFAC, tt}. 

\subsection{Continuous Inputs}

The proposed model so far only works for a discrete input space. Furthermore, it does not imply any smoothness on the values of the input variables. Although, this makes it a powerful, highly non-linear model, it is prone to overfitting. If the input values follow a natural ordering, or if they are discretized from a continuous scale, the model requires many more training examples to learn the smoothness implicitly. To introduce smoothness explicitly, and to extend the model to continuous inputs, we use smooth basis functions for the latent parameters of the decomposition. Instead of indexing the latent representation from a matrix, their values are computed using basis functions. For example, all $A_i$ in equation \ref{parafac} can be modeled using a radial basis function
\begin{equation}
A_i(v_i, r_{i}) = \exp{(- \gamma_{r_{i}} \Vert\mu_{r_{i}} - v_i\Vert^2)}.
\label{rbf}
\end{equation} 
This allows for continuous inputs $v_i \in \setR$. The latent representation is now modeled by the similarity of the input to the center of the radial basis function. In this way, similar inputs induce similar representations. The parameters of the basis function are optimized during training, to yield optimal regression results. Also a mixture of discrete and continuous inputs can easily be modeled, by applying the basis functions only to the continuous inputs, and learning representation matrices for the discrete input variables.

It is also possible to group multiple inputs together into one tensor mode, such that $v_i \in \setR^m$, where $m \in \setN$ denotes the number of grouped inputs. In this way, the representation of a tensor mode is calculated given a vector of continous inputs. The grouping of input variables reduces the dimensionality of the tensor decomposition, and thus the number of free parameters.

\section{Application to Inverse Dynamics}
\label{inverse_dynamics_model}

In the following we describe how the continuous tensor decomposition proposed in section \ref{basis_functions} can be applied to inverse dynamics modeling. 

\subsection{Functional Tucker Decomposition}

We describe a functional Tucker model for the approximation of the joint torques, necessary to perform a movement of a robot arm. Figure \ref{model} shows the model schematically. We consider a robot with $C \in \setN$ degrees of freedom (DoF). In the following we denote the vectors $p, \dot{p}, \ddot{p}$, describing the desired positions, velocities, and accelerations for each of the $c$ DoFs, as $x_1, x_2, x_3 \in \setR^c$ for syntactic reasons. The vector $y \in \setR^c$ describes the corresponding joint torques. 

We model the function $y = f(x_1, x_2, x_3)$ using a functional tensor decomposition model. Each input vector is modeled by one dimension in the tensor decomposition, resulting in third-order tensors $\tensor{Y}$, which describe the joint torques. Each element of the vector $y$ is modeled in a separate model. The resulting three-dimensional tensors of the form \textit{positions} $\times$ \textit{velocities} $\times$ \textit{accelerations}, are then factorized using the Tucker decomposition with limited rank, resulting in a tensor $\Phi \approx \tensor{Y}$, such that
\begin{equation}
\begin{split}
 \Phi(x_1, x_2, x_3) = \sum_{r_1, r_2, r_3}^{\tilde{r}} \tensor{G}(r_1, r_2, r_3) \cdot A_1(x_1, r_1) \cdot A_2(x_2, r_2) \\ 
\cdot A_3(x_3, r_3).
\end{split}
\label{tucker}
\end{equation}
$A_1$ to $A_3$ are functions, which map from the $c$-dimensional input to the latent representations of the Tucker model. We model the representations using multivariate Gaussian kernels, such that
\begin{equation}
\begin{split}
A_i(x_i, r_i) = \exp{\left(-(\mu_{r_{i}} - x_i)^T  D_{r_i} \hspace{0.2em} (\mu_{r_{i}} - x_i)\right)} \hspace{1em} \\
 \forall i \in \{1, 2, 3\},
\end{split}
\end{equation}
with $\mu_{r_{i}} \in \setR^c$ representing the centers and $D_{r_i} \in \setR^{c \times c}$ weighting the distance from the centers in the $c$ dimensional input space. The closer a data point is to the center of a basis function, the higher is its activation. Thus, the centers of the basis functions can be seen as landmarks in the input space. All three-way interactions, between the representations of the three input dimensions, are explicitly modeled and weighted by the elements of the core tensor $\tensor{G}$.

\begin{table*}
	\caption{Normalized mean squared error for all 7 degrees of freedom in percent. Mean and standard deviation of ten random data splits.}
	\begin{center}
		\begin{tabular}{ l || c | c | c | c | c | c | c || c }
			\hline
			Method & DoF 1 &  DoF 2 &  DoF 3 &  DoF 4 &  DoF 5 &  DoF 6 &  DoF 7 & Mean $\pm$ std in \%\\
			\hline
			Linear Regression & 6.80 & 11.62 & 10.82 & 5.81  & 12.81 & 22.59 &  6.73 & 11.03 $\pm$ 0.26 \\ 
			RBF-Network Regression &  2.64 & 1.79 & 1.01 & 0.41 & 4.07 & 3.91 & 1.17 & 2.14 $\pm$ 0.19\\
			Support Vector Regression & 0.88 & 0.67 & 0.43 & 0.15 & 1.04 & 0.72 & 0.34 & 0.60 $\pm$ 0.28 \\
			\hline
			Functional-Tucker & 0.59 & 0.28 & 0.46 & 0.24 & 1.03 & 0.91 & 0.31 & 0.55 $\pm$ 0.24 \\
			Functional-PARAFAC & 1.64 & 1.14 & 0.61 & 0.32 & 1.30 & 1.17 & 0.50 &  0.96 $\pm$ 0.22
		\end{tabular} 
	\end{center}
	\label{table:results}
\end{table*}

\subsection{Model Training}

For training the model, we take a maximum likelihood approach. We minimize the negative log-likelihood of the collected dataset  $\{y^j, (x_1^j, x_2^j, x_3^j)\}_{j=1}^N$ as
\begin{equation}
l = - log \sum\limits_{j=1}^{N} p(y^j | x_1^j, x_2^j, x_3^j, \Theta),
\end{equation}
where $\Theta$ includes the parameters of the decomposition and the basis functions. Assuming a Gaussian distribution, we get the squared error cost function
\begin{equation}
C = \sum_{i=1}^{N} (\tensor{Y}(x_1^j, x_2^j, x_3^j) - \Phi(x_1^j, x_2^j, x_3^j))^2.
\label{costfunction}
\end{equation}
Note, that the cost function considers only nonzero elements of the tensor, i.e., the sparsity of the tensor is exploited. We minimize equation \ref{costfunction} using gradient descent. In experiments, we found the stochastic optimization algorithm Adam, see \cite{adam}, which adopts the learning rate automatically for each parameter, to work best for this task. The sampling of stochastic mini-batches for each update has also shown advantageous, for speeding up training.


We initialize the centers of the Gaussian kernel in a preprocessing step, using three k-means clusterings, such that
\begin{equation}
J_i = \sum_{i}^{\tilde{r}} \sum_{j=1}^{N} \Vert x_i^j - \mu_{r_i} \Vert^2
\end{equation}
are minimized for $i \in \{1, \dots, 3\}$, see \cite{kmeans}. All matrices $D$ are initialized with the identity matrix. The elements of the core tensor $\tensor{G}$ are initialized randomly with a Gaussian distribution of mean zero and standard deviation 0.05. While training all parameters are further optimized. We implemented the model using the computational python library Theano, see \cite{Theano}.

\section{Related Work}
\label{related_work}
Multiway data analysis has found applications in a number of different areas, such as signal processing, neuroscience, and data mining, see \cite{cichocki, morup, PARAFAC, kolda}. Recently, tensor models also have found applications in control engineering such as for modeling hybrid systems, see \cite{lichtenberg1} and multilinear dynamical systems, see \cite{russell}. Furthermore, tensor methods have been applied to Boolean networks, see \cite{boolean} and pneumatic models, see \cite{pneumatic}.

The factorization of sparse matrices has become popular in recommendation systems, especially due to its success in the Netflix challenge, see \cite{netflix}. Extensions to higher order tensors can be found in the modeling of large knowledge bases, such as Yago, DBpedia, or Freebase, see \cite{rescal, nickel2015review}. The multi-graphs have been modeled using sparse three-dimensional tensors and decompositions such as RESCAL, see \cite{rescal, nickel2015review}. The approach of factorizing sparse tensors has further been exploited in \cite{mfi}. Here, the decomposition of sparse tensors is applied to multi-class classification with discrete input features.

Tensor regression methods are concerned with the regression of high dimensional data, structured in a multidimensional array. Tensor methods allow for efficient modeling where traditional methods are often insufficient, due to the complex structure of the data and the high input dimensionality. Tensor regression learns a linear mapping and deals with dense input tensors. Thus, their approach is fundamentally different from ours; see \cite{tr1, tr2}.

Our approach shows some similarities to RBF-networks which are able to approximate any non-linear function  by using radial basis functions. RBF-networks have been successfully applied to a number of tasks including control engineering, see \cite{rbf_nets}. The main difference to our proposed functional Tucker model is that RBF-networks learn one latent representation for the complete input, and map it to the output; whereas, the functional Tucker model learns a representation for each tensor mode and jointes them using the tensor decomposition model. In this way multi-way interactions are modeled explicitly.

Inverse dynamics are traditionally modeled using the rigid-body formulation, see \cite{craig}. However, general regression techniques such as locally weighted projection regression (LWPR), Gaussian Processes, and RBF-networks, have shown advantageous for learning inverse dynamics, see \cite{lwpr, rasmussen}. The topic was subject to a number of studies, see \cite{comparison1, schoelkopf}. Support vector regression has shown superior performance for this task.


\section{Experiments}
\label{experiments}

In this section we evaluate our proposed method on an inverse dynamics dataset including movements from a seven degrees of freedom SARCOS robot arm. We compare against various other state-of-the-art regression techniques for this task.

\subsection{Dataset}
The dataset was introduced by \cite{lwpr}.\footnote{http://www.gaussianprocess.org/gpml/data/} It contains data from a SARCOS robot arm with seven degrees of freedom.  The data was collected from the moving robot arm at 100Hz and corresponds to 7.5 minutes of movement. The dataset includes 21 input dimensions, consisting of 7 joint torques, 7 joint positions, 7 joint velocities, and 7 joint accelerations. The whole dataset consists of 42482 samples. We split the dataset randomly into 90 percent training and 10 percent test data. Additional 5 percent of the training set where used as a validation set. The task is to learn a model on the training data, which models the 7 joint torques, given the positions, velocities and accelerations. The offline learned model can then be applied in the forward controller of the robot. The dataset has been subject to some studies on the topic, see \cite{lwpr, rasmussen}. The regression task has been found to be highly non-linear. Non-linear regression techniques outperformed the rigid-body dynamics formulation by a large margin. The performance of the regression techniques is evaluated on the test set, which includes unseen movements. We repeated the random split 10 times and report the average results and the standard deviation of multiple trials.

\subsection{Baselines}

We compare our model against various state-of-the art regression techniques, modeling the function $y = f(q, \dot{q}, \ddot{q})$. The baseline models we consider are linear regression, RBF-networks and support vector regression. In previous studies support vector regression has shown the best results on this task. For all baseline models a concatenated vector $x = \left[ q, \dot{q}, \ddot{q} \right]$ is built. The linear regression model learns a linear mapping from the inputs to the outputs, such that
\begin{equation}
y = W x + b.
\end{equation}
RBF-networks model the regression problem as,
\begin{equation}
	y = \sum\limits_{i=1}^{\tilde{r}} w_i \exp{ ( -\beta_i \Vert x - c_i \Vert^2 ) }.
\end{equation}
The parameters $c_i$, $\beta_i$ and $w_i$ are learned using backpropagation. We initialized the parameters $c_i$ with the centroids of a k-means clustering on the training data, where $\tilde{r}$ is the number of centroids.

Support vector regression (see \cite{svr}) has shown state-of-the-art results in modeling inverse dynamics. It predicts $y$ as,
\begin{equation}
	y = \sum_{j=1}^{N} (\alpha_j - \alpha_j^\star) k(x_j, x) + b
\end{equation}
with $k(x, x')$ being a kernel function. In the experiments we use a Gaussian kernel. $\alpha_j$ and $\alpha_j^\star$ are Lagrange multipliers, which are determined during optimization. In our experiments we use the libsvm library, see \cite{libsvm}.

Furthermore, we compare the functional Tucker model proposed in Section \ref{inverse_dynamics_model} with a functional PARAFAC model. For the functional PARAFAC model we replace the tensor decomposition in equation \ref{tucker} with a PARAFAC decomposition, as shown in equation \ref{parafac}.

\begin{figure}[!t]
	\scalebox{1}{
		\includegraphics[width=0.5\textwidth]{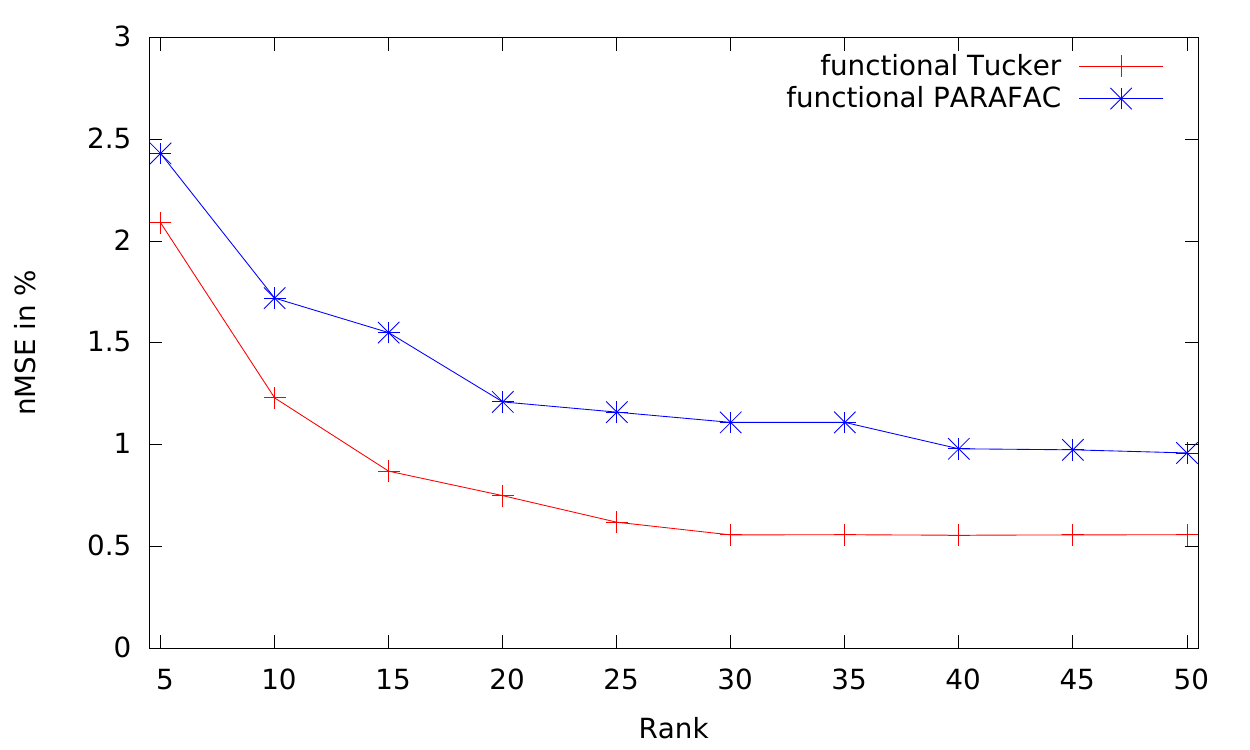}
	}
	\caption{Normalized mean squared error of the functional Tucker and functional PARAFAC model, in dependency of the embedding rank.}
	\label{figure_rank}
\end{figure}

\subsection{Results}

We report the normalized mean squared error (nMSE) for the regression task, which is defined as the mean squared error of all data points divided by the variance of the target variable in the training data. Table \ref{table:results} summarizes the mean nMSE for all seven degrees of freedom in percent. In the rightmost column the mean of all seven degrees of freedom is shown. All results, as well as the standard deviation are referring to the average result of 10 random data splits. The performance of the regression techniques varies across the DoFs. The linear model reaches an nMSE of 11.03\% in average. The nonlinear RBF-networks performs much better with an nMSE of 2.14\% in average. The number of of hidden neurons for the RBF-network was set to 1000. With larger numbers the predictive performance did not increase. The support vector regression model yields a very good result of 0.60\%. Here, we set the parameter C to 600 and $\epsilon$ to 0.1. All hyperparameters were evaluated on a separate validation set. Our proposed functional Tucker model resulted in a slightly better nMSE of 0.55\%. Especially, for the first two DoFs the functional Tucker model performs significantly better than support vector regression. For the other DoFs the results of support vector regression and functional Tucker decomposition are very close to each other.  The parameter efficient functional PARAFAC model reaches an nMSE of 0.96\% in average. Figure \ref{figure_rank} shows the performance of the two functional tensor decomposition models in dependence of the rank of the decompositions. For the Tucker model, the performance converges at a rank of 30 and for the PARAFAC model at a rank of 40. It is also notable that both methods already perform relatively well with a very small rank of 5. The nMSE of the Tucker model is 2.09\% with a rank of 5 and the nMSE of the PARAFAC model is 2.43\%. Both functional tensor models show clearly better results than RBF-networks. This indicates that the explicit modeling of the three-way interaction, yields a significant improvement. 


\section{Conclusion}
\label{conclusion}

In this paper we apply a tensor model, that is based on the Tucker decomposition, to inverse dynamics. Our proposed model exploits the inherent three-way interaction of \textit{positions} $\times$ \textit{velocities} $\times$ \textit{accelerations}. We show how the decomposition of sparse tensors can be applied to regression tasks. Furthermore, we propose to augment the tensor decompositions with basis functions for allowing continuous input variables. In this way, a functional version of a tensor decomposition can be derived. Representations for each tensor mode are induced through the basis functions and fused by the tensor model. The parameters of the basis functions are learned using backpropagation. Experiments on an inverse dynamics dataset, derived from a seven degrees of freedom robot arm, show promising results of our proposed model for the application of learning inverse dynamics. The proposed functional Tucker model outperforms RBF-networks, and even support vector regression, which has shown state-of-the-art performance on this task. Our extension of tensor decomposition models to continuous inputs enables a wide range of application areas. Especially if an inherent multi-way structure exists in the data, functional tensor models can be advantageous over traditional techniques, by explicitly modeling the multi-way interaction. Within automatic robot control the approach might be further extended to learning also a functional Tucker model for a feedback controller based on the tracking errors.

\bibliography{ifacconf}             

\end{document}

%% file: commands.tex
\usepackage{bm}
\newcommand{\tensor}[1]{\ensuremath{\mathcal{#1}}}

\newcommand{\setN}{\ensuremath{\mathbb{N}}}
\newcommand{\setR}{\ensuremath{\mathbb{R}}}